# A New Multifractal-based Deep Learning Model

# for Text Mining

**Zhenhua Wang[1], Ming Ren[1], Dong Gao[2]**

[1]School of Information Resource Management, Renmin University of China, Beijing, China

[2] College of Information Science and Technology, Beijing University of Chemical Technology, Beijing, China

zhenhua.wang@ruc.edu.cn; renm@ruc.edu.cn; gaodong@mail.buct.edu.cn

*Abstract:* In this world full of uncertainty, where the fabric of existence weaves patterns of complexity, multifractal emerges as beacons of insight, illuminating them. As we delve into the realm of text mining that underpins various natural language processing applications and powers a range of intelligent services, we recognize that behind the veil of text lies a manifestation of human thought and cognition, intricately intertwined with the complexities. Building upon the foundation of perceiving text as a complex system, this study embarks on a journey to unravel the hidden treasures within, armed with the proposed multifractal method that deciphers the multifractal attributes embedded within the text landscape. This endeavor culminates in the birth of our novel model, which also harnesses the power of the proposed activation function to facilitate nonlinear information transmission within its neural network architecture. The success on experiments anchored in real-world technical reports covering the extraction of technical term and classification of hazard events, stands as a testament to our endeavors. This research venture not only expands our understanding of text mining but also opens new horizons for knowledge discovery across various domains.

## 1. INTRODUCTION

Text mining aims to automatically and efficiently uncover and explore valuable information or patterns from noisy, irregular and unstructured texts [1-2], thereby enabling us to gain a deeper understanding of the underlying meaning and context within the text, and easily exploring the knowledge, uncovering hidden insights. It can generate informed understanding of the content and has become significant in decision-making in various sectors and domains across industries. For example, we can understand users' preferences [3], sentiments [4], opinions [5], concerns [6] and interests [7] etc., by mining the text generated by users, thus infer their intentions and purposes [8-11]. We are also amenable to the attainment of more sophisticated security risk management practices [12]. Additionally, text mining is responsible for various natural language processing applications such as knowledge graph [13-14], question-answer dialogue system [15-16], and recommendation system [17-18].

Text mining mainly approaches entity recognition and text classification, both of which exhibit certain distinctions in their form. The purpose of entity recognition is to automatically identify expected knowledge from text [19], such as defect knowledge and technical terms in technical reports [20-21]. When it is presented with a sequence of text, $T = <c_1, c_2, \ldots, c_n>$, entities in the form of $<I_b, I_e, \omega, \lambda>$ are recognized. Here, $\omega$ denotes to a particular entity, whereas $I_b$ and $I_e$ are the indexes that highlight the span from $I_b$ to $I_e$, thus specifying the location of the entity within the sequence. Furthermore, $\lambda$ represents the type of the entity in question [22]. For text classification, text $T$ is assigned a category label, such as "*acceptable*, *tolerable*, *investigated*, and *corrected*" for risk evaluation [23], etc. [24-30]

Initial text mining researches predominantly relied on rules. Yet it has been proven to be inadequate when adapting to other projects. Subsequently, researches shifted towards traditional machine learning-based approaches such as support vector machines and decision trees. In pursuit of greater efficiency, supervised deep learning models have emerged. Presently, deep learning has emerged as the paradigm that can learn intricate hidden representations without the need for complex feature engineering. This is accomplished through the text vectorization covering feature capture (such as semantic features derived from BERT.), as well as the design of various neural network architectures (such Attention and BiLSTM), which deliver state-of-the-art performance [19].

Unfortunately, the realm of text mining has largely overlooked the intricate interplay between human cognition, thought processes, and linguistic expressions. In this study, we propose a promising idea that examines text through the lens of fractal analysis, treating it as a complex system. This novel perspective has the potential to unleash unexplored possibilities and invigorate the potential of existing models. The content of text, being influenced by the nuanced choices of words and sentence structures, often exhibits volatile noise patterns. Previous multifractal methods, such as the recent multifractal detrended weighted average algorithm of historical volatility (MF-DHV) [31], have struggled to effectively address this challenge. To alleviate such perturbations, we propose a new multifractal approach termed FS-MFA, which skillfully employs Fourier series to mitigate the inherent noise. By harnessing FS-MFA, we can artfully characterize the intricate features of text, employing the Hurst exponent as a potent descriptor, and harness the carrier of deep learning.

By embracing this paradigm, we can architect a text mining model that astutely captures the multifractal of text using FS-MFA, while leveraging the prowess of neural networks for predictive tasks. Yet, as we advance our models, we acknowledge that prevalent activation functions used in text mining - Tanh, Sigmoid, and ReLU [3-8] - are not exempt from limitations. While each possesses unique strengths, Sigmoid and Tanh encounter challenges related to gradient vanishing, while ReLU confronts the issue of "dead neurons" resulting from zero outputs for negative values [32]. These constraints impede the performance and evolutionary potential of our models. Considering this, we design a new activation function termed Sital by integrating the smoothing characteristics of Tanh and Sigmoid, as well as the positive boundedness of ReLU. Sital disengages gradient vanishing and no-negative values, that can undertake more advanced services.

Formally introduced in this research, a novel deep learning model, DeFFSi, emerges as a groundbreaking thought for text mining, capitalizing on the Sital activation function and FS-MFA framework. DeFFSi's underlying mission revolves around the intricate transformation of text into the vector representations through the utilization of BERT, while concurrently harnessing the informative Hurst exponent profiles generated by FS-MFA. The essence of DeFFSi lies in its ability to seamlessly fuse the semantic richness of text vectors with the multifractal attributes encapsulated within the Hurst exponent series, facilitated by a proposed sophisticated neural network architecture, thereby exhibiting its versatility in addressing diverse text mining challenges. To thoroughly evaluate the efficacy of DeFFSi, a comprehensive case study is meticulously conducted within a realistic context, followed by meticulous experiments targeting technical term recognition and hazard event classification. The empirical findings serve as a resounding testament to the remarkable feasibility of DeFFSi, showcasing the pioneering advancements ushered in by the Sital and remarkable effectiveness of FS-MFA.

As the research unfolds, we anticipate that this paper will not only contribute valuable insights to the field but also provide indispensable decision support for practitioners involved in intricate text mining endeavors, thereby elevating the overall value proposition of this rapidly evolving domain. The main contribution of this study is four-fold.

1. We contribute a new deep learning framework termed DeFFSi for text mining encompassing entity recognition and text classification.

2. We observe the text as a complex system and conceive a new multifractal approach, FS-MFA, to mitigate the noise through Fourier series.

3. We project a new activation function Sital. It can eliminate issues such as gradient vanishing, and stands out from existing competitors

4. We conduct extensive experiments on real-world report mining, thus revealing the effectiveness of DeFFSi, FS-MFA and Sital.

The remainder is organized as follows. Section 2 introduces related work on multifractal analysis and activation function. Section 3 presents DeFFSi in details covering Sital and FS-MFA. Section 4 reports the experiments and their results on the scenarios of technical report. Our research is discussed in Section 5 and concluded in Section 6.

## 2. *RELATED WORK*

### 2.1. Multifractal analysis

Multifractal analysis is a cutting-edge theory that relies on the capricious geometric shapes to characterize complex systems that exhibit self-similarity across a spectrum of scales [33]. It explores the behavior of systems that display varying degrees of variability or irregularity at different scales, achieved by scrutinizing the distribution of the system's fluctuations or variations across different scales. In doing so, it extracts valuable information regarding the system's underlying structure and dynamics. Leveraging the omnipresent fractals in the natural and social sciences offers a unique opportunity to gain profound insights into the underlying mechanisms of complex systems. Currently, this theory is thoroughly ingrained in numerous fields and tasks. The scope of its vitality is not constrained to solely customer comment sentiment analysis, image encryption, gear fault diagnosis, bond assessment, temperature calibration, EEG cognition, artificial translation, and cryptographic systems [34-43].

Multifractal analysis methods aim to investigate the latent fractal attributes present in a range of series. Its procedure typically involves dividing the series data into uniform, small segments, and computing the mean of each segment. Additionally, a polynomial is fitted to the data to eradicate any linear trend, and a fluctuation function is computed for the detrended data. The distribution of the scaling exponents of the fluctuation function is then leveraged to describe the multifractal properties of the series. Popular ones include multifractal detrended fluctuation analysis (MF-DFA) [34], multifractal detrended cross-correlation analysis (MF-DXA) [44] and multifractal detrended weighted average algorithm of historical volatility (MF-DHV) [45]

Within texts, the interplay of words under the influence of grammar creates specific meanings that convey the will, feelings and thoughts of individuals and society. The organization and layout of the text unfold sequentially, adapting in response to shifts in chronological sequencing. Frequently, specific fragments of text can express the meaning of the entire text and may be more precise, succinct, and sophisticated, revealing universal connections in language from a particular vantage point. Hence, text is a complex system that reflects the behavior and thinking behind human cognition. By discovering fractals within text, language itself or certain social behaviors can be more profound and insightful. The Menzerath-Altmann law [46] offers supplementary evidence for the existence of multifractal in languages. It states that as the length of a language structure increases, its components become shorter, implying that the length of a part is a function of the length of the structure. Another reminder is the Hurst exponent [47] reflecting an indication of the autocorrelation of time series, particularly the long-term trends and memories concealed within the series, and is commonly the metric and presentation that portrays the fractal attributes.

An apprehension arises regarding the susceptibility of the meaning conveyed by text to inherent fluctuations caused by the nuanced selection of words and sentence construction. previous multifractal methods may encounter vulnerability in such scenarios. Drawing inspiration from the resplendent efficacy of Fourier transforms in mitigating noise in signal processing [48], we undertake the refinement of MF-DHV and propose FS-MFA, which harnesses the potential of Fourier series to facilitate a novel approach. Hope it to be yielding added incentives for text mining.

## 2.2. Activation function

The indispensability of activation functions is encapsulated in their capacity to instill nonlinearity into deep learning models, and influence the learning trajectory, thereby steering the performance and generalizability of the model. Thus, crafting the optimal activation function tailored to a specific field of research becomes paramount, such as SupEx activation function for detection and classification of pneumonia in CNN [49]. Upon examining activation functions in text mining, Tanh, Sigmoid and ReLU enjoy great popularity

Tanh and Sigmoid (refer to Eq.25-26) is exponential formats, consist of exponentially related terms, such as $e^x$, returning an exponential-conditioned value of the input. They can yield smooth and continuous outputs and directly model complex relationships, thereby seemingly aligning more closely with non-linear mapping relationships.

Nevertheless, their potential downside is that they are susceptible to gradient vanishing as the input value amplifies. Additionally, Sigmoid's output is consistently positive, which may also engender potential issues.

Linear functions, prized for their simplicity and rapid convergence, often find utility within activation functions. The ReLU (refer to Eq.23), a quintessential example, employs a combination of two linear functions with a threshold value of $x = 0$. However, its drawback is the potential for producing "dead" neurons, a condition where the outputs persistently equal zero for all inputs, thus bringing learning to a halt. On the other hand, linear functions commonly serve to temper excessive trends, facilitate the sequential concatenation of piecewise functions, and prevent gradient vanishing, amongst other uses, as illustrated by the likes of SELU [50], Swish [51], and RsigELUD [52].

Promisingly, the notion of designing a new activation function that harnesses the strengths of Tanh, Sigmoid and ReLU could yield considerable potential advantages. By integrating these diverse forms, it's conceivable to construct a more advanced activation function for text mining.

## 3. METHODLOGY

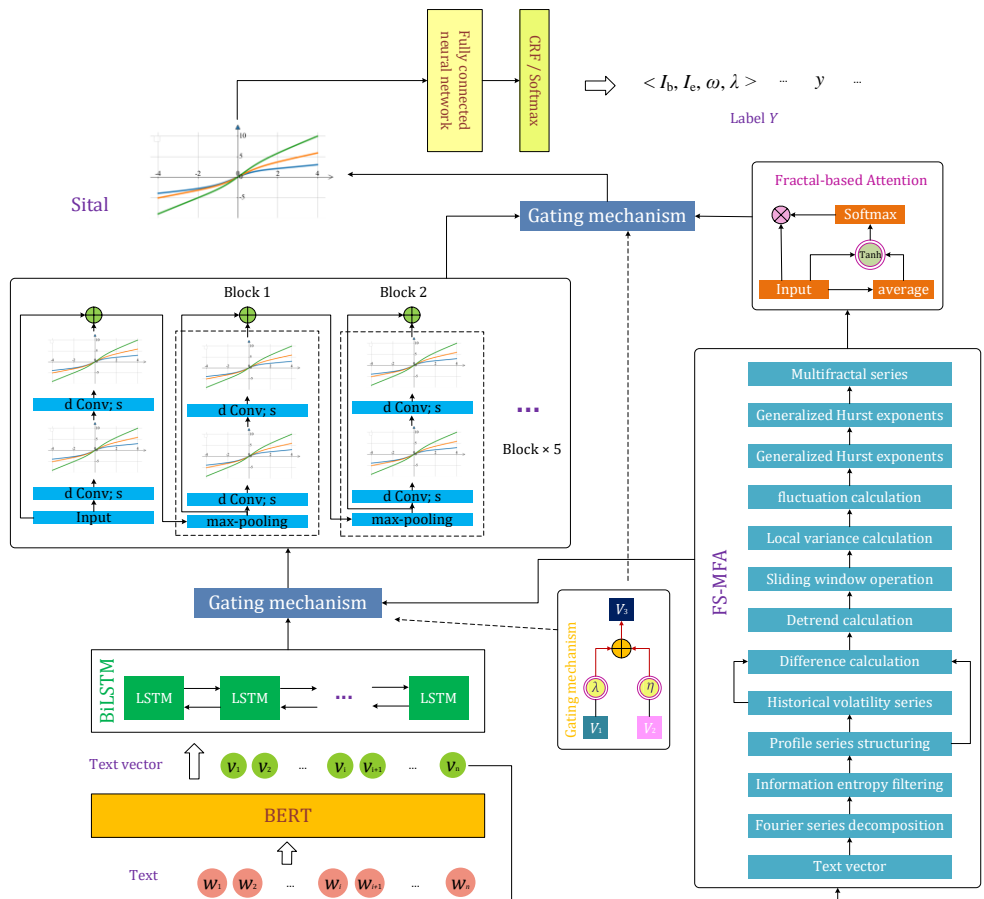

Fig.1: Architecture of the DeFFSi.

We propose DeFFSi, which is a novel deep learning architecture that incorporates multifractal perspectives and achieves more efficient information communication through the utilization of the Sital activation function, as depicted in Fig.1. DeFFSi holds the potential to bring unexpected advancements to the text mining community and serve as a source of inspiration. This section explains DeFFSi. For a comprehensive understanding, we first explicate Sital activation function, followed by a procedure of FS-MFA, and finally dissect the neural network of DeFFSi.

### 3.1. Proposed activation function Sital

Drawing inspiration from the smoothing characteristics of Tanh and Sigmoid, as well as the unbounded nature of ReLU, this study proposes a novel activation function specifically tailored for text mining, termed Sital. By integrating the strengths of these three activation functions, Sital aims to overcome the limitations associated with gradient vanishing and the zero-output dilemma for negative values commonly observed in existing competitors, see Eq.1.

$$Sital(x) = \gamma x + Tanh(x)[1 + \sigma(\eta x)]; \quad Sital\,'(x) = \gamma + [1 - Tanh^2(x)][1 + \sigma(\eta x)] + Tanh(x)[\eta \sigma(\eta x)(1 - \sigma(\eta x))] \quad (1)$$

Where, $\sigma$ is Sigmoid function, $\gamma$ and $\eta$ are learnable parameters that can be fine-tuned and optimized to appropriate values for governing the scaling ranges of the tangent terms on the $y$-axis and $x$-axis, respectively. With Sital(0) = 0, it possesses a zero-centered activation point, akin to Tanh, which exhibits a certain degree of data concentration effect. This characteristic allows Sital to effectively capture and represent the distribution of data by centering the activation around zero, enabling it to better handle the variability and patterns present in the input data.

Fig.2 depict that Sital varies with different parameter settings, i.e., Sital ($\gamma$, $\eta$). In its entirety, Sital exhibits a stable and smooth curve that preserves the characteristic S-shaped style, addressing the limitation of zero output for negative values. By maintaining a continuous and differentiable curve, Sital ensures the smooth flow of information throughout the neural network, enabling it to capture and propagate both positive and negative activations.

The second component of Sital takes the form of the product of Tanh and Sigmoid, and its fluctuation range is unable to surpass the linear form of the first component of Sital. As a result, Sital remains unsaturated throughout its entire activation range, effectively alleviating the issue of gradient vanishing. This characteristic can also be observed from the derivative of Sital, which exhibits a non-zero value across its entire range. By preventing the gradients from approaching zero, Sital ensures the smooth propagation of gradients during backpropagation, enabling effective learning and optimization within the neural network.

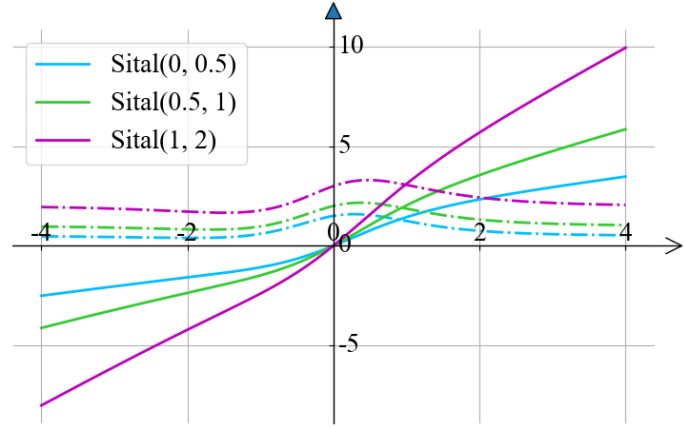

Fig.2: Profile (solid line) and derivative (dashed line) of Sital.

Sital takes on various forms, and its derivative is not a fixed constant, thereby satisfying nonlinearity. It upholds that neural networks refuse to degenerate into a linear transformation with poor generalizability.

Furthermore, unlike ReLU, which is non-differentiable at x=0, Sital is differentiable across its entire domain. This property facilitates gradient-based optimization algorithms, allowing for smoother training of neural networks. Sital utilizes backpropagation for adaptive optimization, and its update way follows the chain rule, see Eq.2. Where, $\psi$ covers parameters $\gamma$ and $\eta$, $C$ is loss function, $f(y_i)$ is the function value of the neuron.

$$\frac{\partial C}{\partial \psi} = \sum_{y_i} \frac{\partial C}{\partial Sital(y_i)} \frac{\partial Sital(y_i)}{\partial \psi} \qquad (2)$$

Throughout the neural network, the parameters $\psi$ can be updated using the Adam optimizer. These parameters are initialized to 1. To prevent the gradient explosion of parameter during the learning process, it is recommended to use a small learning rate. In this study, the learning rate is set to 0.0005.

### 3.2. Proposed Fourier series based multifractal analysis

To mitigate the noise caused by the differences in human cognition reflected in the writing and engraving of text, a novel approach called Fourier Series based Multifractal Analysis (FS-MFA) is proposed. FS-MFA aims to address this issue by retaining high-information components. The details are as follows:

Given a text vector $V = \{v_1, v_2, \ldots, v_n\}$ with length $N$, its Fourier series can be calculated as Eq.3.

$$f(v) = \lim_{r \to \infty} [\eta_0 + \sum_{u=1}^{r} (\alpha_u + \cos(u\omega v) + \beta_u \sin(u\omega v)] \qquad (3)$$

Where, $\eta_0$, $\{\alpha_u\}$ and $\{\beta_u\}$ are the Fourier coefficients, $\omega = 2\pi / T$ is the angular frequency that corresponds to the times that the value of the $V$ goes back and forth between the positive interval and the negative interval, see Eq.4

$$T = \sum_{i=1}^{n-1} \tau; \ \tau = \begin{cases} 1, & if \ v_i v_{i+1} < 0 \\ 0, & if \ v_i v_{i+1} > 0 \end{cases} \qquad (4)$$

The parameter $r$ in the Fourier series represents the number of truncated terms. On one hand, a smaller value of $r$ corresponds to a simpler representation of $f(v)$ using the truncated Fourier series, but it may result in the loss of significant information. On the other hand, using a higher order $f(v)$ with a larger $r$ can lead to insufficient noise mitigation. Therefore, to strike a balance, we consider the concept of information entropy, see Eq.5.

Information entropy is a measure of the average amount of information contained in a random variable. In the context of FS-MFA, we utilize information entropy as a criterion to determine the optimal value of $r$. By calculating the information entropy of the truncated Fourier series representation at different values of $r$, we can assess the value of $r$. This approach allows us to strike a balance between the simplicity of the representation and the preservation of significant information. By employing information entropy as a guiding principle, we aim to achieve an effective and efficient representation of the text data, where the amount of retained information is optimized to mitigate noise and capture the essential characteristics of the text.

$$I(u) = -\sum_{u=1}^{r} p_i \log_2(p_i) \qquad (5)$$

Where, $p_i$ represents the frequency of each term in $V$. By analyzing the information entropy at different values of $r$, we observe the changes in $I(u)$. The value of $r$ that corresponds to the point where $I(u)$ experiences the smallest drop is selected as the order to retain in the Fourier series, see Eq.6, $I'(u)$ is the derivative of $I(u)$.

$$r = u; \ if \ I'(u) \cdot I'(u+1) < 0 \qquad (6)$$

Therefore, $V$ is denoised in this form as $\tilde{V}$, see Eq.7.

$$f(\overset{\sim}{v}) = \eta_0 + \sum_{u=1}^{R} (\alpha_u + \cos(u\omega v) + \beta_u \sin(u\omega v) \qquad (7)$$

For $\tilde{V} = \{\tilde{v}_1, \tilde{v}_2, ..., \tilde{v}_n\}$, we structure its profile series $Y$ as Eq.8.

$$Y = \sum_{i=1}^{k} (\tilde{v_i} - \frac{1}{n} \sum_{i=1}^{n} \tilde{v_i}), \ k = 1, 2, ..., n \qquad (8)$$

We calculate the historical volatility series of $Y$, see Eq.9. Where, $\theta$ is the historical volatility of the log sequence of series.

$$y_i^{'} = \frac{\sum_{t=1}^{s} \theta_{i-t}}{\sum_{t=0}^{m} \theta_{i-t}} y_{i-1}^{'} + \frac{\theta_i}{\sum_{t=0}^{m} \theta_{i-t}} Y_i; \ \ y_1^{'} = \frac{1}{s} \sum_{i=s}^{2s-1} Y_i \qquad (9)$$

We can obtain the detrend series $D$ of $Y$ through Eq.10.

$$D(k) = y_k^{'} - Y_k; \ \ 2s \le k \le N \qquad (10)$$

Further, we process the $D$ in a sliding window manner $W_s = [N / s]$, where, $[\cdot]$ is rounding function, $s$ is the window scale. Accordingly, obtain residual series at $m$-th window, see Eq.11.

$$D_m(k) = D(W_s + k); \ \ 1 \le k \le 2s, \ l = ms \qquad (11)$$

We can reveal the long-range correlation of $Y$ under its representation and behavior through alleviating its non-stationarity. Eq.12 calculates the local variance $\sigma^2(s, m)$ controlled by the $s$ variable (window scale) and $m$ variable (time window series number).

$$\sigma^2(s, m) = \frac{1}{s} \sum_{k=1}^{s} (D_m(k))^2 \qquad (12)$$

For $W_s$ windows, Eq.13 calculates the mean value of $\sigma^2(s, m)$ to reap the $q$-order fluctuation function $F_q(s)$ [52].

$$F_q(s) = \begin{cases} [\frac{1}{W_s} \sum_{m=1}^{W_s} \sigma^2(s,m)^{\frac{q}{2}}]^{\frac{1}{q}}, \ q \ne 0 \\ exp\{\frac{1}{2W_s} \sum_{m=1}^{W_s} ln(\sigma^2(s,m))\}, \ q = 0 \end{cases} \qquad (13)$$

We observe $F_q(s)$ by setting variable $s$, obviously, with the increase of $s$, $F_q(s)$ rises in a power-law expressed in Eq.14 which depends on the fractal order $q$.

$$F_q(s) \propto s^{H(q)} \qquad (14)$$

We leverage Boltzmann method [53] to fit $F_q(s)$, and can obtain the FS-MFA-conditioned generalized Hurst exponents $H(q)$ ($H(q)$ is the standard one under $q = 2$), which forms the multifractal vector series of text with multifractal particularities that are reflected in the variation of $H(q)$ with $q$.

### 3.3. Proposed DeFFSi

With great anticipation, this section elucidates the neural network embodied by DeFFSi. Formally, given a text $W$ formulated as a word sequence $\{w_1, w_2, …, w_n\}$ with the length $N$, DeFFSi aims to extract the expected entities from it and to classify it into appropriate category in the form of labels $Y$.

DeFFSi first calls prevalent BERT that enjoys semantic feature capture to embed $W$ to form a vector $V = \{v_1, v_2, …, v_n\}$, where, $V \in R^{d \times n}$, $d$ represents the dimension of $V$. Subsequently, $V$ undergoes two journey-like of processing.

1. $V$ is fed into two bidirectional LSTMs to generate hidden vectors $H^{(t)} = \{h_1^{(t)}, h_2^{(t)}, …, h_n^{(t)}\}$, where $t \in \{0, 1, 2, …, T\}$. The $H^{(0)}$ is obtained through initialization. Each current hidden state $h_i^{(t)}$ is a combination of the current input vector $v_i$, as well as the future state $h_{i+1}^{(t)}$ and past hidden state $h_{i-1}^{(t)}$, see Eq.15. As a result, $H^{(t)}$ captures contextual features.

$$h_i^{(t)} = [LSTM\,(v_i, h_{i-1}^{(t)}); LSTM\,(v_i, h_{i+1}^{(t)})] \qquad (15)$$

2. $V$ is averaged and transformed into the multifractal vector $FV$ through FS-MFA.

Next, $H^{(t)}$ and $FV$ are fused across a gating mechanism $G$, where $G$ is formulated using Sigmoid function producing continuous outputs bounded between 0 and 1, see Eq.16 [54], $\kappa$ and $b$ are the weight matrix and bias term, and $\lambda$ and $\eta$ are the input weights satisfying $\lambda + \eta = 1$, here i.e., for $H^{(t)}$ and FV, respectively, and $FVH$ is the fusion vector.

$$\lambda = Sigmoid(\kappa \cdot H + b)$$
$$V_3 = \lambda \cdot V_1 + \eta \cdot V_2 \qquad (16)$$

Then, $FVH$ is transmitted to the proposed CNN architecture with Sital Activation function, namely SCNN，to diligently unearth profound local features. SCNN consists of an input pre-processing module followed by five identical blocks in its architectural structure. The input pre-processing module takes the $FVH$ as its input and generates a feature graph $p_1$ through the convolution kernels with the $d$ ($d = 4$) width and $s$ ($s = 256$) filters. With the blessing of Sital, the feature map $p_1$ is then nonlinearly propagated to the next convolutional operation, generating the deeper-level feature map $p_2$, see Eq.17. Where, $FVH_{k:\,k+d-1}$ is a fragment of $FVH$ indexes from $k$ to $k+d$-1, $n$ is the length of $FVH$. Following the principle of residual operations, the concatenation of $p_2$ and *FVH* is employed as the input to block1. Transitioning to block1, the input undergoes a *max*-pooling operation to accentuate the primary feature vectors *FVH-$p_2$*. Subsequently, *FVH-$p_2$* undergoes two rounds of convolution operations, same as the previous steps, resulting in the generation of feature map $p_4$. In block2, the concatenation of $p_4$ and *FVH-$p_2$* serves as the input, which is further accentuated through *max*-pooling to form the feature vector *FVH-$p_4$*, see Eq.18. Continuing the iterative process, the SCNN yields the concatenated representation of $p_{12}$ and *FVH-$p_{10}$*, resulting in the emergence of the highly sophisticated *FVHC* vector.

$$p_i = \{Sital(W_{p_{i-1}} \cdot FVH_{k:k+d-1} + b_{p_{i-1}})\}_m \mid_{m=1}^{m=n-d+1} \qquad (17)$$

$$FVH\text{-}p_i = \max\text{-}pooling(FVH_{p_{i-2}} \oplus p_i) \qquad (18)$$

Meanwhile, the *FV* traverses an attention mechanism based on additive scoring and soft-alignment functions, striving to capture the multifractal features. Notably, the *FV*, along with its mean $\overline{FV}$, is employed to compute attention scores using Eq.19, subsequently generating attention weight matrices through the *Softmax*, see Eq.20

$$\delta = Q^T Tanh(W^{\overline{FV}} \overline{FV} + W^{FV} FV + b) \qquad (19)$$

$$a_i = \frac{\exp(\delta_i)}{\sum_{i=1}^{n} \exp(\delta_i)} \qquad (20)$$

Where, $Q^T$, $W^{\overline{FV}}$ and $W^{FV}$ are the trained weight matrices, $b$ is the bias term。 The attention output *FVA* of *FV* is reflected in Eq.21.

$$FVA_i = a_i FV_i \qquad (21)$$

Following the acquisition of the feature vectors *FVHC* and *FVA*, a gating mechanism $G$ is employed to blend their representations. This fusion is subsequently channeled through a cascading sequence of fully connected neural networks, adorned with the Sital activation function. Finally, employ either the *Softmax* / CRF for predictions in the classification / entity recognition.

## 4. *EXPERIMENTS*

This study introduces a case of technical report mining, corresponding to technical term recognition and hazard event classification involved respectively.

## 4.1. Data description

The technical report employed is HAZard and OPerability analysis (HAZOP) reports. As a risk assessment approach that has become the de facto industry standard, HAZOP strives to spot, appraise, and manage deficiencies in various industry and social systems. It takes the deviation of nodes within the systems as the entry to investigate and analyze its underlying causes, consequences, suggestions and measures, and the results are documented in the HAZOP report [54]. As HAZOP reports embody invaluable expert experience and procedural knowledge, their mining can offer vital support for sustainable development of industry. HAZOP reports involve loads of entities, which are technical terms under unique working conditions. They can be categorized as equipment (e.g., Fischer-Tropsch reactor), materials (e.g., hexamethylene diisocyanate), consequences (e.g., high liquid level) and state (e.g., insufficient oxygen content). Recognizing these technical terms can provide support for many promising applications in practice, for example, it can build an industry safety knowledge graph which greatly supports industry examining [12]. HAZOP reports are a repository of rich and insightful information about hazard events, and classifying them is conducive to the overall planning and decision-making for industry development [55]. For instance, Fang et al., developed a classification model to assist experts and engineers in decision making and hidden hazard treatment respectively [23]. The hazard events can be measured in terms of severity, possibility and risk. Regarding severity, i.e., the degree of damage such as personal injury, hazard events can be classified into level 1 to 5, indicating *negligible*, *minor*, *serious*, *disabled*, and *death* respectively. In terms of possibility, i.e., the frequency of triggering, hazard events can be classified into level 1 to 5, indicating that the occurrence of hazard events is *none*, *10 years ago*, *within 10 years*, *within 5 years*, and *within 2 years* respectively. On risk, hazard events can be classified into level 1 to 4 under the acceptability of existing preventive measures, indicating that it is *acceptable*, *tolerable*, *investigated*, and *corrected*. The following text describes a hazard event about E-5611102 (a kind of circulating heat exchanger separator), and it is classified as {4, 1, 2} in terms of severity, possibility and risk.

" Due to the high temperature of light oil and gas from E-5611102, the temperature of the incoming gas from the circulating gas is high, and the heavy oil flows into the subsequent process, resulting in the blockage of the pipeline of the air cooler AE-5611101, and the evacuation and damage of the circulating gas compressor. Carry out the interlock protection of the compressor itself. It is recommended to clarify the treatment means and guidance requirements of the compressor. "

## 4.2. Tasks

### 4.2.1. Technical term recognition

There are two datasets termed GSSL and CPSYB for technical term recognition. Table 1 shows the annotation of technical terms. CPSYB and GSSL have 78315 and 72189 rows of data respectively, and are randomly divided into training set, test set and validation set in the ratio of 8:1:1.

Table 1: Annotation of technical terms.

| Term category | Initial | Subsequent |
|---|---|---|
| Equipment | B-EQU | I-EQU |
| Consequence | B-CON | I-CON |
| Material | B-MAT | I-MAT |
| State | B-STA | I-STA |
| No term | O | O |

### 4.2.2. Hazard event classification

Through text preprocessing, 5869 hazard events and their labels are collected as datasets of severity, possibility and risk. See Table 2 for the quantity of each level. They are randomly assigned to training set, test set and validation set in the ratio of 8:1:1.

Table 2: Hazard event classification dataset.

| Theme | Level #1 | Level #2 | Level #3 | Level #4 | Level #5 |
|---|---|---|---|---|---|
| Possibility | 419 | 1760 | 1607 | 1134 | 949 |
| Severity | 1570 | 2732 | 1353 | 170 | 44 |
| Risk | 2902 | 2577 | 335 | 55 | - |

### 4.3. Experiment #1: Evaluate the performance of DeFFSi

Regarding technical term recognition, three frequently-used models are built for comparison [12, 22]. BiLSTM-CNN-CRF, BERT-BiLSTM-CRF and BERT-CNN-BiLSTM-CRF. In terms of hazard event classification, BERT-RAtt, BERT-DPCNN and BERT-RCNN serve as comparative trials because their popularity [23, 56-57].

### 4.4. Experiment #2: Evaluate the performance of Sital

Based on DeFFSi model, the performance of the proposed Sital activation function is evaluated. In each round of experiments, except for the replacement of activation functions, the model remained stationary. The commonly used activation functions are baselines, including GELU [58], ReLU, Leaky ReLU, Sigmoid, Tanh, ELU [59], SELU [50], Softplus [60], Swish [51], RsigELUD ($\alpha_1 = 0.05$, $\alpha_2 = 0.2$) [52] and KDAC [19], as listed in Eq.22-32 respectively. Sigmoid, Tanh and ReLU rule text mining, and Leaky ReLU is a well-known variant of ReLU. KDAC is committed to text knowledge discovery. GELU is the built-in activation function of BERT. The other five activation functions have achieved the success in the computer vision tasks, and are also included for the comparison.

$$GELU(x) = \frac{1}{2}x(1 + \tanh(\sqrt{2/\pi}(x + 0.044715x^3))) \qquad (22)$$

$$ReLU(x) = \begin{cases} 0, x < 0 \\ x, x \geq 0 \end{cases} \qquad (23)$$

$$Leaky \ \ ReLU(x) = \begin{cases} \alpha x, x < 0 \\ x, x \geq 0 \end{cases} \qquad (24)$$

$$Sigmoid(x) = \frac{1}{1 + e^x} \qquad (25)$$

$$Tanh(x) = \frac{e^x - e^{-x}}{e^x + e^{-x}} \qquad (26)$$

$$ELU(x) = \begin{cases} \alpha(e^x - 1), x < 0 \\ x, x \geq 0 \end{cases} \qquad (27)$$

$$SELU(x) = \lambda \begin{cases} \alpha(e^x - 1), x < 0 \\ x, x \geq 0 \end{cases} \qquad (28)$$

$$Softplus(x) = \ln(1 + e^x) \qquad (29)$$

$$Swish(x) = \frac{x}{1 + e^{\beta x}} \qquad (30)$$

$$RSigELUD(x) = \begin{cases} \alpha(x/(1 + e^{-x})) + x, x > 1 \\ x, 0 < x < 1 \\ \beta(e^x - 1), x < 0 \end{cases} \qquad (31)$$

$$KDAC(x) = \begin{cases} P_{Max}(P_{Min}(Tanh, \beta_1 x), \beta_2 x) \\ P_{Min}(f_i, f_j) = f_i + \varsigma(f_j - f_i) + \mu\varsigma^2 - \mu\varsigma \ ; \varsigma = \frac{1}{2} + \frac{f_i - f_j}{2\mu}, \xi = \frac{1}{2} + \frac{f_j - f_i}{2\mu} \\ P_{Max}(f_i, f_j) = f_i + \xi(f_j - f_i) - \mu\xi^2 + \mu\xi \end{cases} \qquad (32)$$

### 4.5. Experiment #3: Evaluate the effectiveness of FS-MFA

To assess the effectiveness of FS-MFA, it is juxtaposed with the widely employed MF-DFA [34], MF-DHV [45]. MF-DXA [44] as comparative benchmarks within the framework of DeFFSi. In each evaluation, except for the replacement of multifractal methods, the model remained stationary.

### 4.6. Experiment setting

The parameters in each round of experiments are consistent. Specifically, the optimizer is Adam, the learning rate is 0.0003, BERT with a size of base that meets the default parameters, the epoch is 20, and the num of hidden units in LSTM is 200. The result is the average of five repeated experiments. The metric is F1.

## 5. RESULTS

### 5.1. Results #1: performance of DeFFSi

This section gives the results of evaluating DeFFSi, see Table 3-4, where "test" and "val" refers to the test set and the validation set respectively. In terms of recognizing technical terms, DeFFSi demonstrates superior performance, surpassing its competitors, particularly the BiLSTM-CNN-CRF, with a significant performance improvement of approximately 7% and 4% on the test and validation sets of GSSL and CPSYB, respectively. Furthermore, DeFFSi outperforms the other two competitors by about 1% on each dataset. Switching to hazard event classification, DeFFSi is equally impressive. In comparison to BERT-DPCNN, DeFFSi exhibits a performance advantage of roughly 4% on both the test and validation sets for severity classification, and about 3% on the validation set for risk classification. Hence, DeFFSi not only excels in technical term recognition, but also exhibits remarkable proficiency in hazard event classification. It is undoubtedly competent for text mining.

Table 3: Evaluation results of technical term recognition.

| Models / Datasets | GSSL | | CPSYB | |
|---|---|---|---|---|
| | test | val | test | val |
| BiLSTM-CNN-CRF | 62.18 | 34.28 | 82.94 | 82.41 |
| BERT-BiLSTM-CRF | 68.03 | 40.33 | 84.78 | 85.35 |
| BERT-CNN-BiLSTM-CRF | 67.82 | 40.07 | 85.27 | 85.64 |
| DeFFSi | **69.34** | **41.52** | **86.83** | **86.39** |

Table 4: Evaluation results of hazard event classification.

| Models / Datasets | Severity | | Possibility | | Risk | |
|---|---|---|---|---|---|---|
| | test | val | test | val | test | val |
| BERT-RAtt | 77.38 | 77.21 | 70.11 | 70.08 | 70.75 | 70.99 |
| BERT-DPCNN | 75.54 | 76.58 | 70.20 | 69.98 | 70.84 | 69.97 |
| BERT-RCNN | 77.26 | 78.92 | 70.95 | 70.16 | 70.93 | 70.13 |
| DeFFSi | **79.64** | **80.37** | **72.11** | **71.50** | **71.89** | **72.76** |

### 5.2. Results #2: profit of Sital

Table 5: Evaluation results of activation function.

| Activation function | GSSL | | CPSYB | | Severity | | Possibility | | Risk | |
|---|---|---|---|---|---|---|---|---|---|---|
| | test | val | test | test | val | val | test | val | test | val |
| GELU | 67.80 | 41.20 | 85.70 | 85.70 | 79.38 | 79.91 | 71.50 | 69.76 | 71.74 | 71.34 |
| ReLU | 69.17 | 41.14 | **87.08** | 86.19 | 79.19 | 79.40 | 70.41 | 71.25 | 69.96 | 72.25 |
| Leaky ReLU | 68.53 | 39.68 | 85.37 | 85.49 | 78.79 | 80.09 | 70.95 | **71.61** | 71.01 | 71.98 |
| Sigmoid | 68.39 | 41.13 | 85.99 | 86.02 | 78.25 | 79.41 | 70.67 | 70.55 | 71.41 | 71.48 |
| Tanh | 67.71 | 40.08 | 84.88 | 86.21 | 79.28 | 78.80 | 70.33 | 70.85 | 71.20 | 71.55 |
| ELU | 68.13 | 40.22 | 86.13 | 85.71 | 78.50 | 79.53 | 71.60 | 70.89 | 71.49 | 70.94 |
| SELU | 68.65 | 39.60 | 86.64 | 85.18 | 79.42 | 79.58 | 71.83 | 70.00 | 70.96 | 71.61 |
| Softplus | 69.02 | 39.84 | 86.58 | 86.06 | 79.20 | 79.99 | 71.23 | 71.17 | 70.55 | 71.63 |
| Swish | 68.99 | 40.32 | 86.26 | 85.33 | 79.38 | 80.06 | 70.42 | 69.54 | 70.37 | 72.26 |
| RSigELUD | 68.65 | 39.69 | 86.03 | 85.28 | 79.03 | 78.49 | 71.87 | 71.09 | 69.89 | 71.21 |
| KDAC | 67.55 | 39.87 | 85.06 | 85.82 | 79.18 | 80.29 | 71.67 | 69.78 | 70.13 | 72.30 |
| Sital | **69.34** | **41.52** | 86.83 | **86.39** | **79.64** | **80.37** | **72.11** | 71.50 | **71.89** | **72.76** |

This section reports the results of evaluating the activation function, as displayed in Table 5. Out of the ten evaluation sets across five datasets on two tasks, Sital achieves the best performance in eight of the sets, indicating its progressiveness and superiority.

Specifically, on GSSL dataset, Sital demonstrates significant superiority over its competitors, occupying all comparisons in both the test and validation sets. Specifically, in validation set, Sital exhibits a lead of 1.5% over KDAC, RSigELUD, Softplus, SELU and Leaky ReLU, and a lead of more than 1% over Swish, ELU, and Tanh. The test set also demonstrates that Sital outperforms its rivals, surpassing GELU etc., by more than 1.5 percentage point, and by approximately 1% in Sigmoid and ELU. The superiority of Sital on CPSYB dataset has also demonstrated, Sital improves by1.5% over Tanh and KDAC in test set. While Sital's performance is slightly inferior to ReLU, this difference is small. In validation set, Sital has an improvement of 1%+ compared to RSigELUD, SELU and Swish, and an approximately 0.7% enhancement over ELU, Leaky ReLU and GELU. Results of severity classification clearly reflects that Sital has an overwhelming advantage against its competitors, especially in comparison to RSigELUD and Tanh on validation set, along with Sigmoid and ELU on test set, with a lead of more than 1.5% in former, and 1% in latter. With outstanding performance Sital more or less surpasss other competitors, despite its underwhelming performance in certain situations such as competing with Swish and Softplus etc. Let's switch our perspective to the results on possibility classification. Sital's performance achieves 71.50% on validation set, which is only 0.11% (Leaky ReLU) away from the first place, see Fig.11. Sital outdistances KDAC, Swish, SELU and GELU by more than 1.5% in test set, and validation set indicates that it has the same superiority confront Swish, Tanh and ReLU. Sital also significantly defeats Sigmoid etc. Regarding results on risk classification. Sital performs the best in both test and validation sets, and surpasses RSigELUD (all sets), ELU (val), KDAC (test), Swish (test) and ReLU (test) by a margin of 1.5+ percentage points. Not to be overlooked, Sital beats Softplus, SELU and Leaky ReLU by a not insignificant margin of about 1%. No state to fall behind competitions

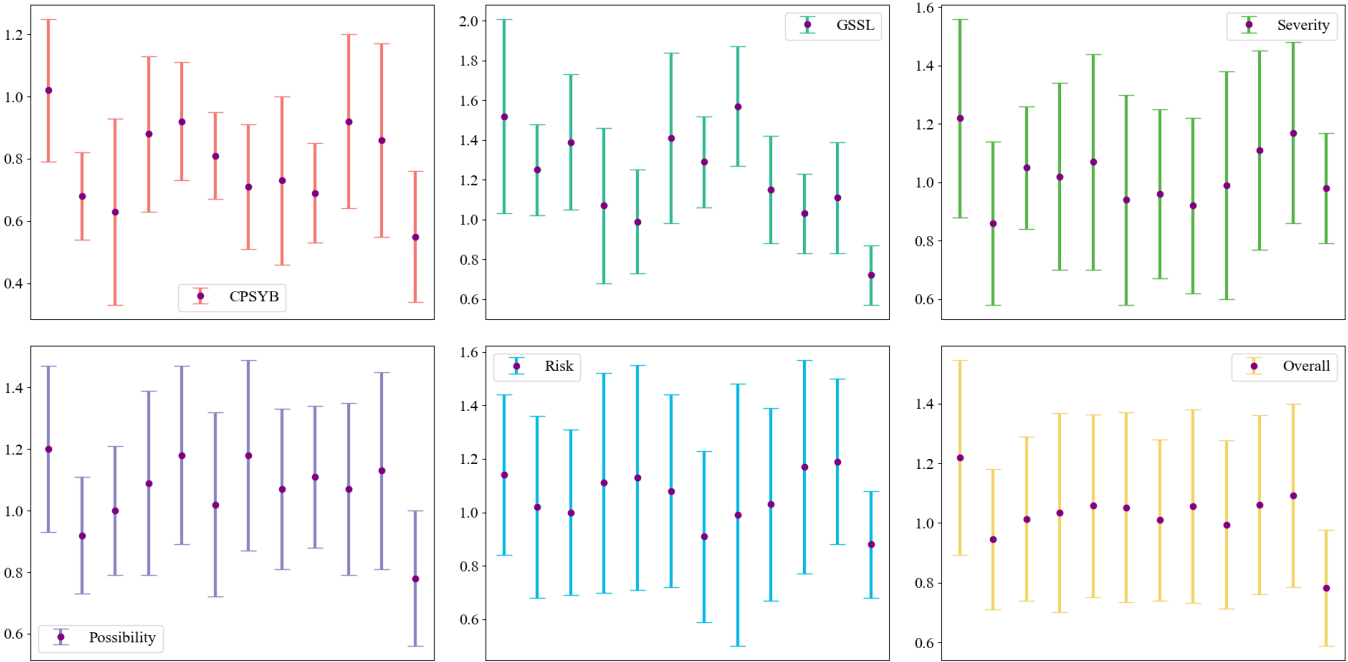

Fig.3: the mean and standard deviation of training loss. In each subplot, GELU, ReLU, Leaky ReLU, Sigmoid, Tanh, ELU, SELU, Softplus, Swish, RsigELUD, KDAC and Sital are presented in sequence.

Fig.3 illustrates the training loss of the model across various tasks (the first five subplots) and the overall (the last subplot) individually under each activation function. Notably, it is discernible that the model trained with the Sital activation function yield remarkably lower loss values, thereby exhibiting a heightened level of competitiveness. This compelling evidence firmly attests to the efficacy of Sital in enabling more optimal and resilient model training, thus engendering enhanced performance and an elevated competitive edge.

## 5.3. Results #3: effectiveness of FS-MFA

Table 6 substantiate the superior effectiveness of the proposed FS-MFA compared to rivals. Notably, with the exception of a marginal 0.14% lag behind MF-DFA in the test set for probability classification, FS-MFA consistently outperforms its counterparts across the remaining nine evaluation sets, indicating its unrivaled prowess. This affirms the remarkable efficacy and versatility of FS-MFA in capturing the intricate multifractal characteristics inherent in the texts, thereby empowering it to excel in a wide range of text mining tasks and consolidate its status as the preferred method of choice.

Table 6: Performance (%) of DeFFSi under different multifractal methods

| Multifractal methods | GSSL | | CPSYB | | Severity | | Possibility | | Risk | |
|---|---|---|---|---|---|---|---|---|---|---|
| | test | val | test | test | val | val | test | val | test | val |
| MF-DXA | 68.53 | 41.26 | 86.50 | 85.96 | 79.05 | 79.89 | 71.97 | 71.45 | 71.46 | 72.28 |
| MF-DHV | 69.02 | 40.89 | 86.61 | 86.11 | 79.42 | 80.28 | 71.84 | 71.33 | 71.79 | 72.57 |
| MF-DFA | 68.85 | 41.07 | 85.94 | 85.83 | 79.17 | 80.02 | **72.25** | 71.14 | 71.68 | 72.55 |
| FS-MFA | **69.34** | **41.52** | **86.83** | **86.39** | **79.64** | **80.37** | 72.11 | **71.50** | **71.89** | **72.76** |

## 6. DISCUSSION

To sum up, the proposed DeFFSi enjoys the mighty power by undergoing extensive experiment evaluations. DeFFSi is shaped under deep learning, supported by Sital, and draws on the FS-MFA that depict text with complex system. Its excitement can be attributed to nonlinear transfer and fitting of Sital, as well as the additional profound stimulation portrayed by FS-MFA.

Sital is a groundbreaking activation function that has emerged as a formidable competitor in its field. Its success can be primarily attributed to its meticulously designed structure, which carefully incorporates the strengths of Tanh, Sigmoid, and ReLU. By seamlessly integrating the characteristics of exponential and linear functions in a unified framework, Sital inherits their merits and overcomes the challenges commonly plagued in previous text mining endeavors, i.e., gradient vanishing and absence of negative values. Furthermore, Sital's efficacy is enhanced by the inclusion of two learnable parameters, allowing it to approach the inherent diversity of textual data with composure and adaptability. Its remarkable performance is a testament to its ability to optimize text mining community.

It is worth highlighting that in comparison to Tanh, Sigmoid, and ReLU, the inclusion of two additional parameters in Sital introduces a heightened level of space complexity to the model. Nonetheless, this increment in complexity can be deemed acceptable, considering that the pursuit of optimal performance frequently necessitates a larger parameter space, as exemplified by the BERT and GPT families. Another reminder is that, the time complexity of Sital can be approximated as $O(e^{3n})$, surpassing that of Tanh ($O(e^{2n})$) and other similar functions. Nevertheless, it remains significantly lower than the complexity of more intricate alternatives such as GELU ($O(ne^{6n})$) and its counterparts. Consequently, Sital maintains a favorable position in terms of time complexity, demonstrating its viability for practical implementations. The added evidence is that text mining is typically performed offline and time-insensitive. The last concern is that the "No Free Lunch Theorem" has proved the rationality of the existence of Sital. Text mining, a scientific research field with broad potential, deserves attention and exploration of its activation function. Sital hopes to become the default of text mining to achieve better service for decision support.

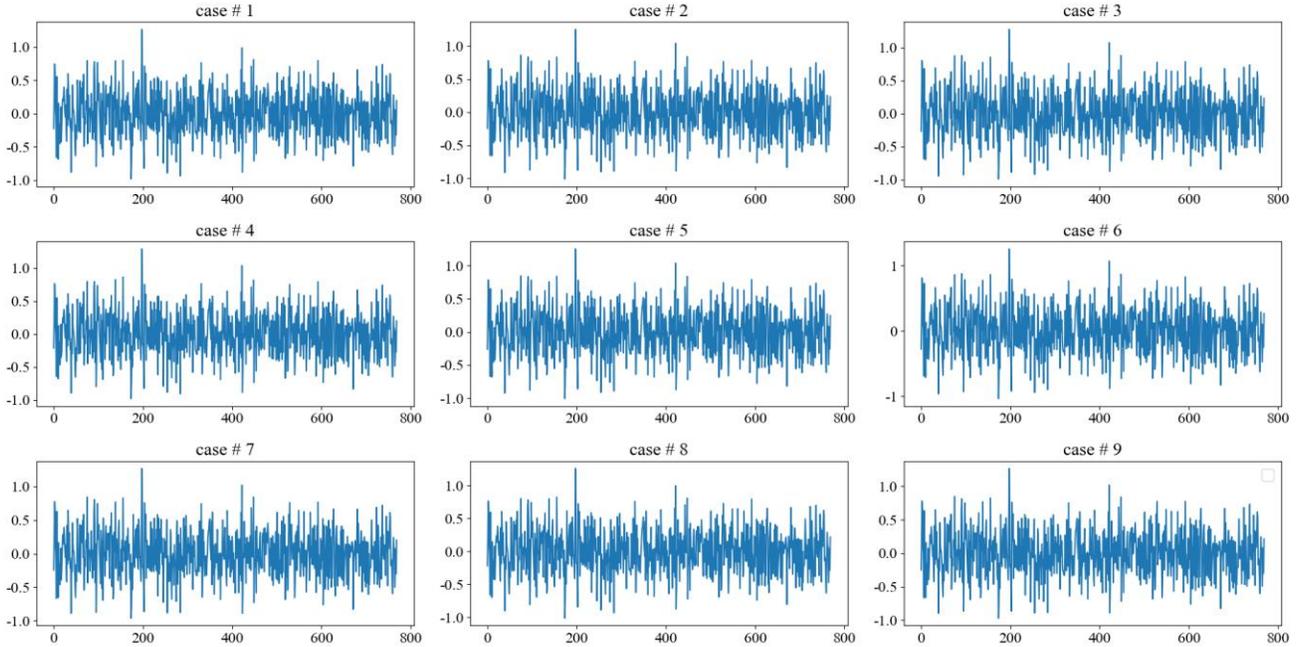

Fig.4: Text vectors on nine description cases.

The perspective vacates to fractals, and our motivation is to view text as the complex system of how words are arranged and combined in human thinking. The fractal theory is used to facilitate a glimpse of the truth, i.e., part of the text is advertised as a whole, thus attempting to feed a novel incentive to the model. We propose FS-MFA to inscribe the multifractal feature series of the text vector and make it run into the model through two nodes, one by fusing it with the output of BiLSTM through a gating mechanism, and the other by coupling it after *Attention* processing with SCNN through a gating mechanism. In these ways, the model is sequentially stimulated. Moreover, the gating mechanism can control their assigned weights, and there is less concern that a certain class of features becomes the landlord or that certain features may be anomalies that pull down the model.

The progressiveness of FS-MFA is confirmed in the experiments. We here present several engaging observations and discussions. To illustrate the multifractal processing, we have randomly chosen nine descriptions, labelled from case #1 through case #9, within technical reports. Fig.4 presents these descriptions, acquired through the application of BERT vectorization. They reveal a high degree of complexity, rendering direct analysis a challenging endeavor.

Fig.5 presents the contour of text vectors after the removal of potential noise (see Eq.7), revealing their transformation from sharp and jagged patterns to relatively smooth shapes. These filtered vectors can be interpreted as mitigating the influence of rhetorical devices and writing styles that manifest in the text, while preserving the core meaning conveyed by the textual content. Consequently, this transformation enables us to gain more convenient insights and decipher the fractal properties inherent, thereby enhancing our ability to comprehend the underlying patterns and structures embedded within the text.

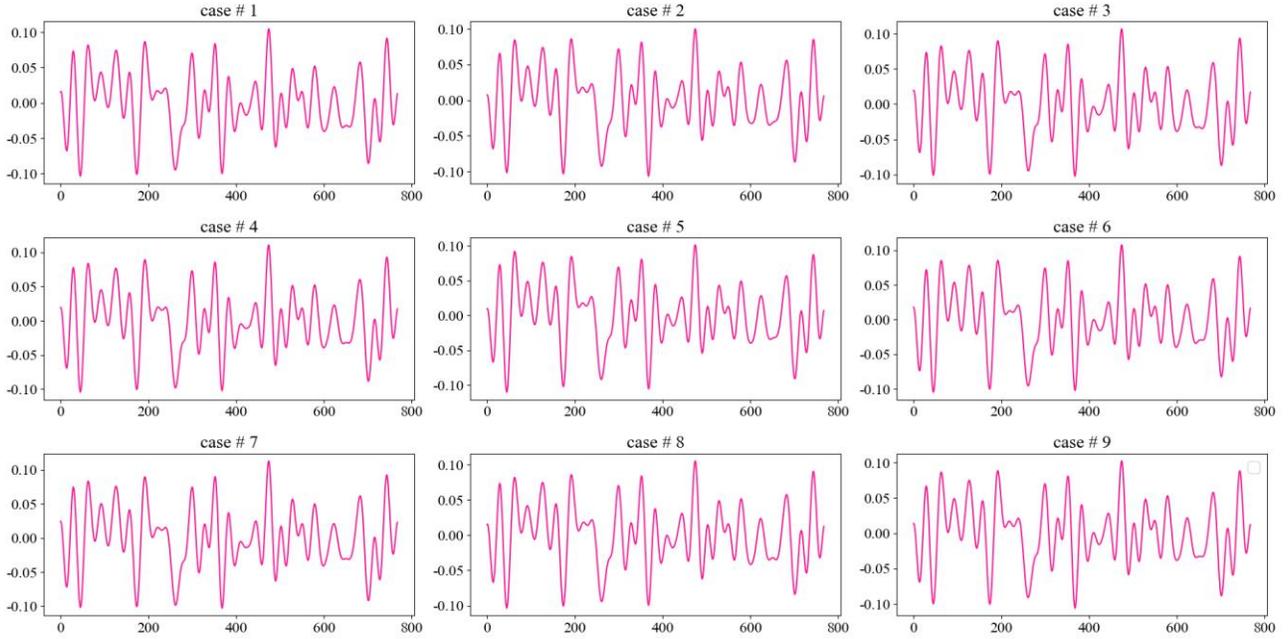

Fig.5: Denoising series conditioned on Fourier series and information entropy.

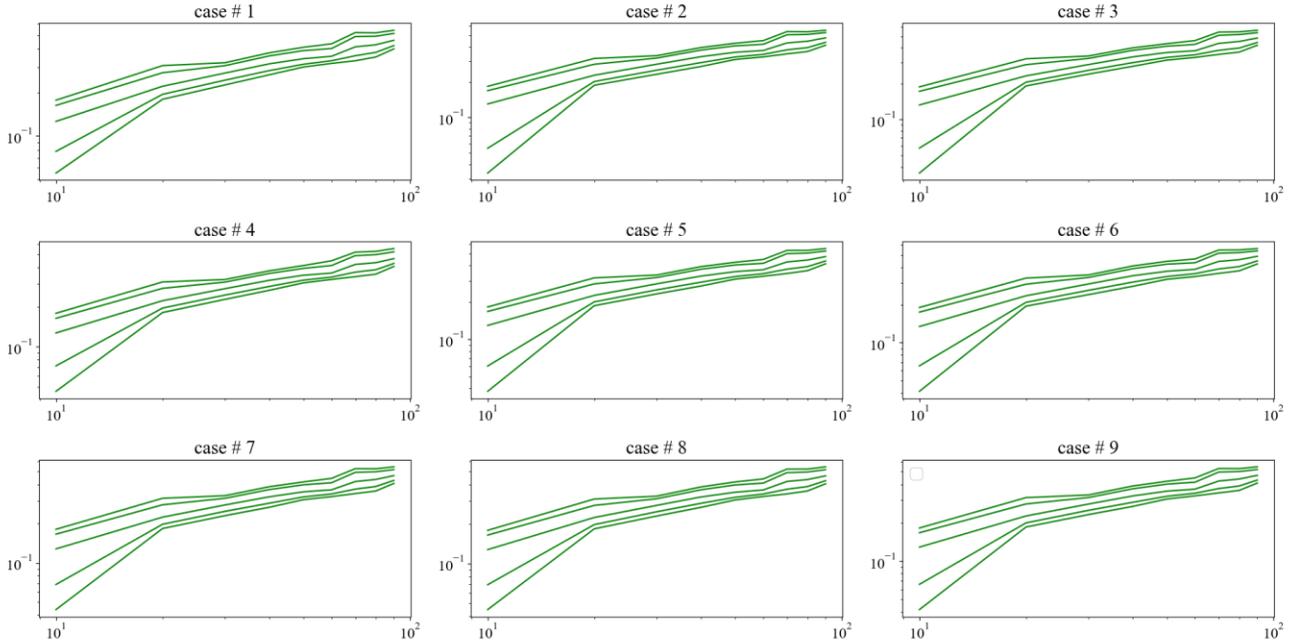

Fig.6: Curves of the fluctuation function $F_q(s)$.

Fig.6 depicts the fluctuation function $F_q(s)$ (see Eq.13) under a logarithmic coordinate system, where the curve from bottom to top represents $q$ = -10, -5, 0, 5, 10. The variable slopes of these curves affirm our ability to characterize their multifractal properties. Moreover, the fluctuation curve's modification in response to varying $q$ values suggests that the features display diverse behaviors dependent on the scale or amplitude of fluctuations. Nonetheless, the overarching trend remains congruent, hinting at a consistent writing theme across the text.

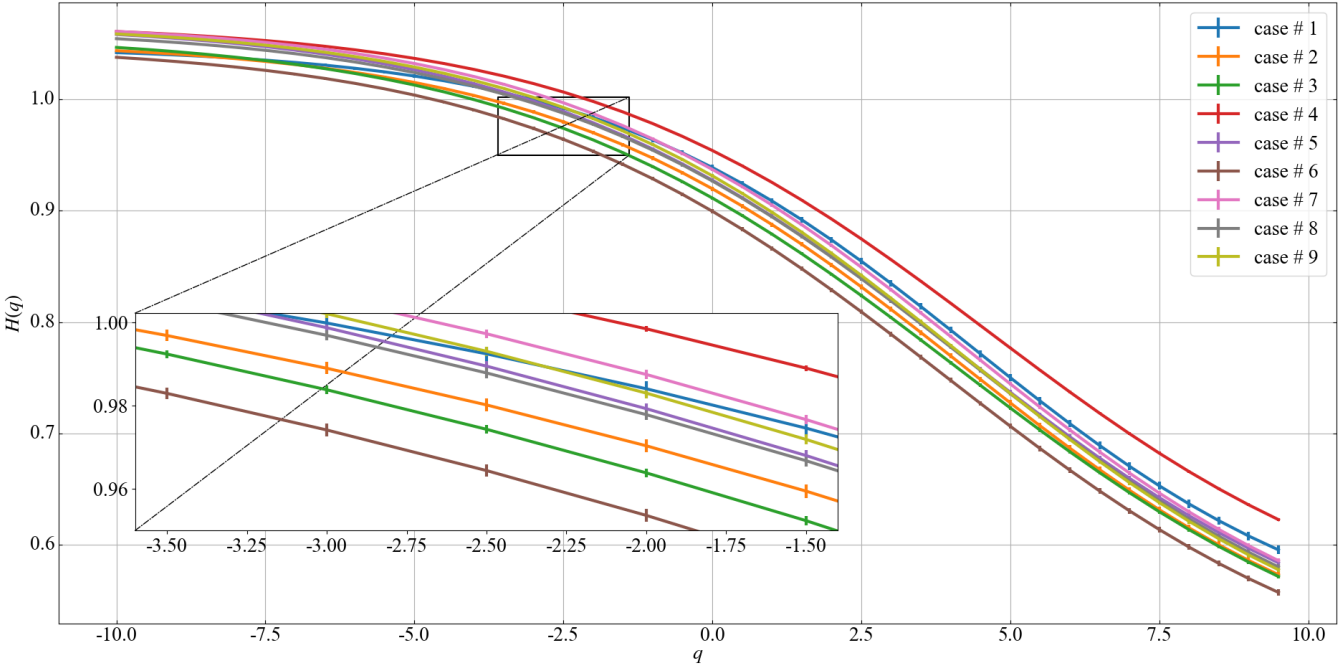

Fig.7: Display of $H(q)$.

Fig.7 presents the varying profiles of the generalized Hurst exponent, $H(q)$, observed across nine cases. The inconsistency in the values of $H(q)$ for these cases signifies the presence of local fluctuations and intersecting points, underscoring the unique fractal characteristics inherent in each case. These individual characteristics can subsequently serve as a guiding force, facilitating the model's ability to accurately classify the categories to which they belong.

The value of $H(q)$ dynamically shifts in response to $q$, implying that the scaling properties are subject to change contingent upon the amplitude of the fluctuation under consideration. This particular observation is emblematic of multifractal behavior, suggesting a nuanced and intricate structure embedded within the technical report. $H(q)$ portrays varying scaling behaviors contingent upon the level of fluctuations. Specifically, a value of H(q) exceeding 0.5 suggests the presence of a long-term persistent trend within the sequence. This implies that as the distance between points broadens, the correlation among these data points progressively diminishes. This behavior demonstrates that elements within the text of a technical report, be they words or phrases, are not isolated entities，rather exhibit intricate interconnections with various sections of the text, which assists in the mining of technical reports when scrutinizing and deducting through the lens of deep learning.

It should be noted that our data originates from a typical and grounded technical report, despite not being openly accessible to be a limitation, does not affect the relevance and practicality of our findings. Also, it has practical value and meets realistic needs and can be acceptable.

With utmost sincerity, we ardently aspire that our research endeavors will yield auspicious outcomes, propelling the provision of invaluable services to the public.

## 7. CONCLUSION

In culmination, we have unveiled a new research in text mining through our novel model, DeFFSi. By assimilating the multifractal stimulation harnessed from FS-MFA and leveraging the information transmission capabilities facilitated by Sital, our research encapsulates the following content.

New deep learning framework: DeFFSi integrates an amalgamation of neural network modules, fostering a fertile terrain for the exploration of text mining and heralding unprecedented possibilities for knowledge discovery.

Innovative multifractal analysis method: FS-MFA, characterized by the filtering mechanism of fusing Fourier series and information entropy, unlocks a new chapter in the realm of multifractal exploration within text, unraveling intricate patterns and unveiling latent structures.

Novel activation function: Sital amalgamates the virtues of Sigmoid, Tanh, and ReLU, the preeminent activation functions in text mining. It overcomes the challenges of gradient vanishing and negative value absence, enabling smoother nonlinear information transmission.

Comprehensive experimental trials: Our extensive suite of experiments conducted on authentic datasets substantiates the robustness and efficacy of DeFFSi, FS-MFA and Sital, as their triumphs across a broad spectrum of real-world scenarios, underscoring their potential as potent tools for practical applications.

Management insights: Our research offers valuable insights into information processing and management, shedding light on the vast untapped potential within text as a complex system. This opens new avenues for harnessing text mining to extract valuable knowledge and support decision-making processes.

However, we must acknowledge the vistas that remain uncharted, beckoning further exploration and refinement. While our research has cast a discerning eye on multifractality in text, there exists an avenue for the granular investigation of specific textual lexicons, alongside their intricate correlations and self-similarity. This dimension of inquiry beckons future investigations to traverse the juncture where multifractal research intersects with the nuanced realm of linguistics, thereby ensuring a more comprehensive comprehension. Moreover, our efforts to mitigate noise have unveiled the influence of literary stylings and rhetorical flourishes on textual content. Alas, the task of localizing such noise within specific textual components persists as an open challenge. Furthermore, an ardent focus on more intricate neural network architectures, such as their synergistic amalgamation with graph neural networks, holds great promise in fostering a harmonious convergence of complexity science and network-centric paradigms.

In summation, our research endeavors aspire to actuate a transformative shift in information processing and management. By unraveling the cryptic dimensions of text as a complex system, we envisage a future where knowledge discovery and decision support transcend conventional boundaries, engendering a paradigm of enlightened information processing and management practices.

## ACKNOWLEDEMENTS

This work is supported by the major program of Renmin University of China (No. 21XNL019), and the National Natural Science Foundation (NNSF) of China under Grants 61703026 and 61873022.